\documentclass[10pt,twocolumn,letterpaper]{article}

\usepackage{iccv}
\usepackage{times}
\usepackage{epsfig}
\usepackage{graphicx}
\usepackage{amsmath}
\usepackage{amssymb}
\usepackage{epstopdf}
\usepackage{amsmath}

\usepackage{graphicx}
\usepackage{makecell}
\usepackage{epstopdf}
\usepackage{amsfonts}
\usepackage{color}
\usepackage{multirow}
\usepackage{setspace}
\usepackage{booktabs}
\usepackage{enumerate}
\usepackage[linesnumbered,ruled,vlined]{algorithm2e}
\def\degree{${}^{\circ}$}

\usepackage[pagebackref=true,breaklinks=true,letterpaper=true,colorlinks,bookmarks=false]{hyperref}

\iccvfinalcopy 

\ificcvfinal \fi
\begin{document}

\title{Multiple Human Association between Top and Horizontal Views by Matching Subjects' Spatial Distributions}

\author{Ruize Han, Yujun Zhang, Wei Feng$^*$, Chenxing Gong, Xiaoyu Zhang, Jiewen Zhao, Liang Wan, Song Wang \\
School of Computer Science and Technology, Tianjin University, Tianjin, China\\ 
}

\maketitle

\begin{abstract}
	
Video surveillance can be significantly enhanced by using both top-view data, e.g., those from drone-mounted cameras in the air, and horizontal-view data, e.g., those from wearable cameras on the ground. Collaborative analysis of different-view data can facilitate various kinds of applications, such as human tracking, person identification, and human activity recognition. However, for such collaborative analysis, the first step is to associate people, referred to as subjects in this paper, across these two views. This is a very challenging problem due to large human-appearance difference between top and horizontal views. In this paper, we present a new approach to address this problem by exploring and matching the subjects' spatial distributions between the two views. More specifically, on the top-view image, we model and match subjects' relative positions to the horizontal-view camera in both views and define a matching cost to decide the actual location of horizontal-view camera and its view angle in the top-view image. We collect a new dataset consisting of top-view and horizontal-view image pairs for performance evaluation and the experimental results show the effectiveness of the proposed method.

\end{abstract}

\section{Introduction}

\begin{figure}[t]
	\centering
	\includegraphics[width=0.475\textwidth]{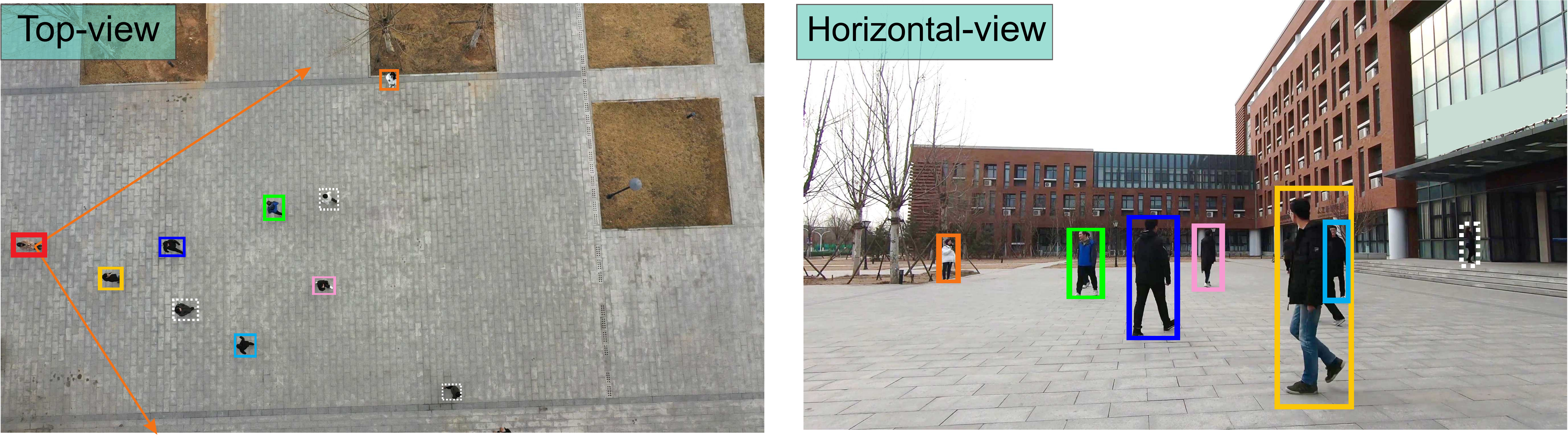}
	\caption{An illustration of the top-view (left) and horizontal-view (right) images. The former is taken by a camera mounted to a drone in the air and the latter is taken by a GoPro worn by a wearer who walked on the ground. The proposed method identifies on the top-view image the location and view angle of the camera (indicated by red box) that produces the horizontal-view image, and associate subjects, indicated by identical color boxes, across these two videos.}
	\label{fig:example}
\end{figure}

The advancement of moving-camera technologies provides a new perspective for video surveillance. Unmanned aerial vehicles (UAVs), such as drones in the air, can provide top views of a group of subjects on the ground. Wearable cameras, such as Google Glass and GoPro, mounted over the head of a wearer (one of the subjects on the ground), can provide horizontal views of the same group of subjects. As shown in Fig.~\ref{fig:example}, the data collected from these two views well complement each other -- top-view images contain no mutual occlusions and well exhibit a global picture and the relative positions of the subjects, while horizontal-view images can capture the detailed appearance, action, and behavior of subjects of interest in a much closer distance.  Clearly, their collaborative analysis can significantly improve the video-surveillance capabilities such as human tracking, human detection, and activity recognition. 

The first step for such a collaborative analysis is to accurately associate the subjects across these two views, i.e., we need to identify any person present in both views and identify his location in both views, as shown in Fig.~\ref{fig:example}. In general, this can be treated as a person re-identification (re-id) problem -- for each subject in one view, re-identify him in the other view. However, this is a very challenging person re-id problem because the same subject may show totally different appearance in top and horizontal views, not to mention that the top view of subjects contains very limited features by only showing the top of heads and shoulders and it can be very difficult to distinguish different subjects from their top views, as shown in Fig.~\ref{fig:example}. 

Prior works~\cite{ardeshir2016ego2top,Ardeshir2018Egocentric,ardeshir2018integrating} tried to alleviate the challenge of this problem by assuming 1) the view direction of the top-view camera in the air has certain slope such that subjects' body, and even part of the background, are still partially visible in top views and can be used for feature matching to the horizontal views, and 2) the view angle of the horizontal-view camera on the ground is consistent with the moving direction of the camera wearer and can be easily estimated by computing optical flow in the top-view videos This can be used to identify the on-the-ground camera wearer in the top-view video. These two assumptions, however, limit the their applicability in practice, e.g., the horizontal-view camera wearer may turn head (and therefore the head-mounted camera) when he walks, leading to inconsistency between his moving direction and wearable-camera view direction.

In this paper, we develop a new approach to associate subjects across top and horizontal views without the above two assumptions. Our main idea is to explore the spatial distribution of the subjects for cross-view subject association. From the horizontal-view image, we detect all the subjects, and estimate their depths and spatial distribution using the sizes and locations of the detected subjects, respectively. On the corresponding top-view image, we traverse each detected subject and possible direction to localize the horizontal-view camera (wearer), as well as its view angle. For each traversed location and direction, we estimate the spatial distribution of all the visible subjects. We finally define a matching cost between the subjects' spatial distributions in top and horizontal views to decide the horizontal-view camera location and view angle, with which we can associate the subjects across the two views. In the experiments, we collect a new dataset consisting of image pairs from top and horizontal-views for performance evaluation. Experimental results verify that the proposed method can effectively associate multiple subjects across top and horizontal views. 

The main contributions of this paper are: 
1) We propose to use the spatial distribution of multiple subjects for associating subjects across top and horizontal views, instead of using subject appearance and motions in prior works. 
2) We develop geometry-based algorithms to model and match the subjects’ spatial distributions across top and horizontal views.
3) We collect a new dataset of top-view and horizontal-view images for evaluating the proposed cross-view subject association. 

The remainder of this paper is organized as follows. Section~\ref{sec:related} reviews the related work. Section~\ref{sec:method} elaborates on the proposed method and Section~\ref{sec:experiment} reports the experimental results, followed by a brief conclusion in Section~\ref{sec:conclusion}.

\section{Related Work}
\label{sec:related}

Our work can be regarded as a problem of associating first-person and third-person cameras, which has been studied by many researchers. For example, Fan et al. \cite{fan2017identifying} identify a first-person camera wearer in a third-person video by incorporating spatial and temporal information from the videos of both cameras. In \cite{ferland2009egocentric}, information from first- and third-person cameras, together with laser range data, is fused to improve depth perception and 3D reconstruction. Park et al. \cite{park2013predicting} predict gaze behavior in social scenes using both first- and third-person cameras. In \cite{xu2018joint}, first- and third-person cameras are synchronized, followed by associating subjects between their videos. In \cite{soran2014action}, a first-person video is combined to multiple third-person videos for more reliable action recognition. The third-person cameras in these methods usually bear horizontal views or views with certain slope angle. Differently, in this paper the third-person camera is mounted on a drone and produces top-view images, making cross-view appearance matching a very difficult problem. 

As mentioned above, cross-view subject association can be treated as a person re-id problem, which has been widely studied in recent years. Most existing re-id methods can be grouped into two classes: similarity learning and representation learning. The former focuses on learning the similarity metric, e.g., the invariant feature learning based models~\cite{Liao2015Person, Rui2014Learning,Yang2014Salient},  classical metric learning models~\cite{K2012Large, Paisitkriangkrai2015Learning, Giuseppe2015Person}, and deep metric learning models~\cite{Low2016Learning, Varior2016A}. The latter focuses on feature learning, including
low-level visual features such as color, shape, and texture ~\cite{Gray2008Viewpoint, Ma2012Local}, and more recent CNN deep features~\cite{Zheng2017Person, Tong2016Learning}.
These methods assume that all the data are taken from horizontal views, with similar or different horizontal view angles, and almost all of these methods are based on appearance matching. In this paper, we attempt to re-identify subjects across top and horizontal views, where appearance matching is not an appropriate choice.

More related to our work is a series of recent works by Ardeshir and Borji~\cite{ardeshir2016ego2top,Ardeshir2018Egocentric,ardeshir2018integrating} on building association between top-view and horizontal-view cameras. In \cite{ardeshir2016ego2top,Ardeshir2018Egocentric}, by jointly handling a set of egocentric (first-person) horizontal-view videos and a top-view video, a graph-matching based algorithm is developed to locate all the horizontal-view camera wearers in the top-view video. In \cite{ardeshir2018integrating}, the problem is extended to locate not only the camera wearers, but also other horizontal-view subjects in the top-view video. However, as mentioned above, these methods are based on two assumptions: 1) the top-view camera bears certain slope angle to enable the partial visibility of human body and the use of appearance matching for cross-view association, and 2) the looking-at direction of the horizontal-view camera is the same as the moving direction of the camera wearer. In this paper, we remove these two assumptions and leverage the spatial distribution of subjects for cross-view subject association. The methods developed in \cite{ardeshir2016ego2top,Ardeshir2018Egocentric,ardeshir2018integrating} require multi-frame video inputs since it needs to estimate each subject's moving direction, while our method can associate a single frame in top view and its corresponding frame in horizontal view.

\section{Proposed Method}
\label{sec:method}
In this section, we first give an overview of the proposed method and then elaborate on the main steps. 

\subsection{Overview}

\begin{figure}[htbp]
	\centering
	\includegraphics[width=\columnwidth]{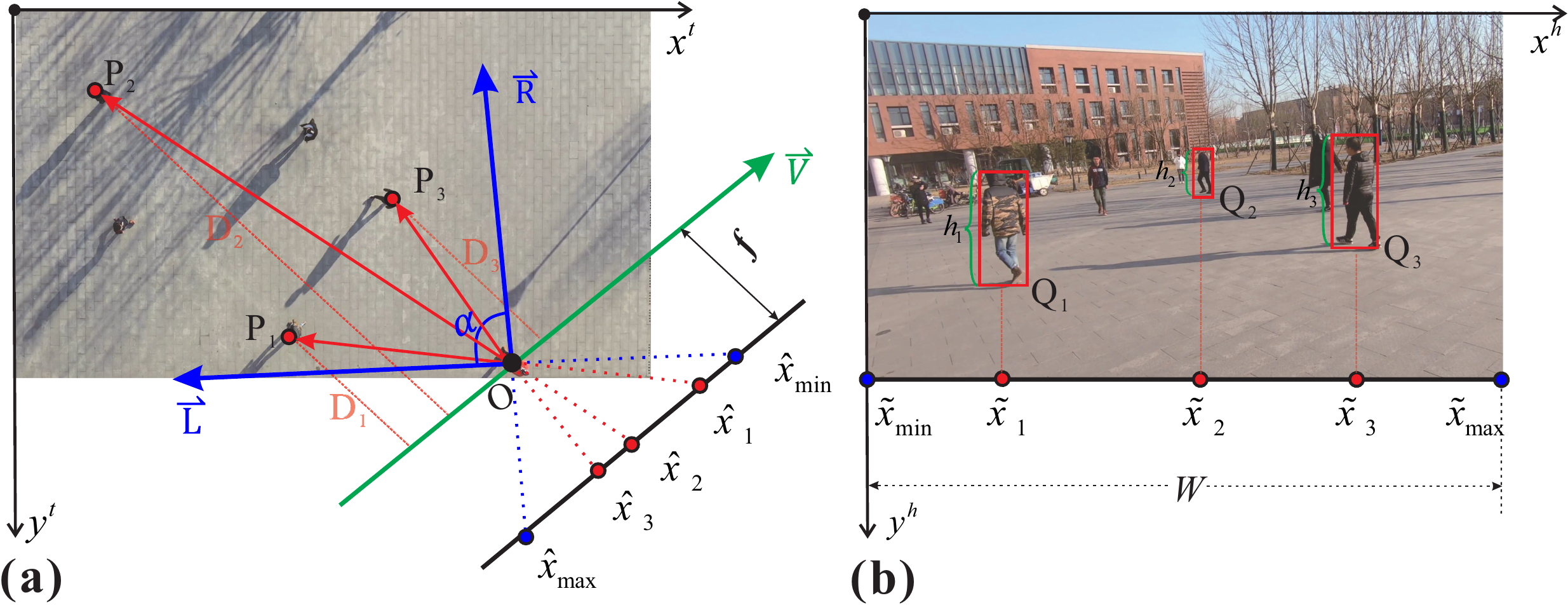}
	\caption{An illustration of vector representation in (a) top view and (b) horizontal view.}
	\label{fig:top_view}
\end{figure}

Given a top-view image and a horizontal-view image that are taken by respective cameras at the same time, we detect all persons (referred to as subjects in this paper) on both images by a person detector~\cite{Redmon2016You}. 
Let ${\mathcal{T}} = \{ O_i^{{\mathop{\rm top}\nolimits} }\}_{i=1}^{N}$ be the collection of $N$ subjects detected on the top-view image, 
with $O_i^{{\rm{top}}}$ being the $i$-$\mathrm{th}$ detected subject. 
Similarly, let ${\mathcal{E}} = \{ O_j^{{\mathop{\rm hor}\nolimits} }\}_{j=1}^M$ be the collection of $M$ subjects detected on the horizontal-view image, with $O_j^{{\rm{hor}}}$ being the $j$-$\mathrm{th}$ detected subject. 
The goal of cross-view subject association is to identify all the matched subjects between ${\mathcal{T}} $ and ${\mathcal{H}}$ that indicate the the same persons.

In this paper, we address this problem by exploring the spatial distributions of the detected subjects in both views. 
More specifically, from each detected subject $O_i^{{\rm{top}}}$ in the top view, we infer a vector $\mathbf{V}^\mathrm{top}_i = (x^\mathrm{top}_i, y^\mathrm{top}_i)$  
that reflects its relative position to the horizontal-view camera (wearer) on the ground. 
Then for each detected subject $O_j^{{\rm{hor}}}$ in the horizontal view, we also infer a vector $\mathbf{V}^\mathrm{hor}_j = (x^\mathrm{hor}_j, y^\mathrm{hor}_j)$ to reflect its relative position to the horizontal-view camera on the ground. 
We associate the subjects detected in two views by seeking matchings between two vector sets 
${{\bf{V}}^{{\mathop{\rm top}\nolimits} }}({\mathcal{T}},\theta,O) = \{\mathbf{V}^\mathrm{top}_i\}_{i=1}^N$
and $\mathbf{V}^\mathrm{hor} ({\mathcal{E}})= \{\mathbf{V}^\mathrm{hor}_j\}_{j=1}^M$, 
where $O$ and $\theta$ are the location and view angle of the horizontal-view camera (wearer) in 
the top-view image and they are unknown priorly. 
Finally, we define a matching cost function $\phi$ to measure the dissimilarity between the two vector sets and optimize this function for finding the matching subjects between two views,
as well as the camera location  $O$, and camera view angle $\theta$. 
In the following, we elaborate on each step of the proposed method. 

\subsection{Vector Representation}

In this section, we discuss how to derive $\mathbf{V}^\mathrm{top}$ and $\mathbf{V}^\mathrm{hor}$. On the top-view image, we first assume that the horizontal-view camera location $O$ and its view angle $\theta$ are given. This way, we can compute its field of view in the top-view image and all the detected subjects' relative positions to the horizontal-view camera on the ground. Horizontal-view image is egocentric and we can compute the detected subjects' relative positions to the camera based on the subjects' sizes and positions on the horizontal-view image.

\subsubsection{Top-View Vector Representation}

As shown in Fig.~\ref{fig:top_view}(a), in the top-view image we can easily compute the left and right boundaries of the field of view of the horizontal-view camera, denoted by $\mathord{\buildrel{\lower3pt\hbox{$\scriptscriptstyle\rightharpoonup$}} \over L}$,
$\mathord{\buildrel{\lower3pt\hbox{$\scriptscriptstyle\rightharpoonup$}} \over R}$, respectively, based on the given camera location  $O$ and its view angle $\theta$.  
For a subject at $P$ in the field of view, we estimate its relative position to the horizontal-view camera by using two geometry parameters ${\hat x}$ and ${\hat y}$, where
${\hat x}$ is the (signed) distance to the horizontal-view camera along the (camera) right direction $\mathord{\buildrel{\lower3pt\hbox{$\scriptscriptstyle\rightharpoonup$}} \over V}$, as shown in 
Fig.~\ref{fig:top_view}(a) and ${\hat y}$ is the depth. 
Based on pinhole camera model, we can calculate them by
\begin{equation} \label{eq:xy_top}
\left\{ \begin{array}{l}
{\hat x} = f\cot \langle \mathord{\buildrel{\lower3pt\hbox{$\scriptscriptstyle\rightharpoonup$}} \over {OP}}, \mathord{\buildrel{\lower3pt\hbox{$\scriptscriptstyle\rightharpoonup$}} 	\over V} \rangle\\
{\hat y} =|\mathord{\buildrel{\lower3pt\hbox{$\scriptscriptstyle\rightharpoonup$}} 	\over {OP}}|\cdot \sin \langle \mathord{\buildrel{\lower3pt\hbox{$\scriptscriptstyle\rightharpoonup$}} 	\over {OP}}, \mathord{\buildrel{\lower3pt\hbox{$\scriptscriptstyle\rightharpoonup$}}  \over V} \rangle, 
\end{array} \right.
\end{equation}
where $ \langle \cdot, \cdot \rangle$ indicates the angle between two directions and $f$ is the focus length of horizontal-view camera. 

Next we consider the range of $\hat x$. From Fig.~\ref{fig:top_view}(a), we can get
\begin{equation} \label{eq:xlr}
\left\{ \begin{array}{l}
{\hat x}_{\mathrm{min}} = f \cot  \langle \mathord{\buildrel{\lower3pt\hbox{$\scriptscriptstyle\rightharpoonup$}} \over {L}}, \mathord{\buildrel{\lower3pt\hbox{$\scriptscriptstyle\rightharpoonup$}} 	\over V} \rangle  = f\cot (\frac{{\pi  + \alpha }}{2})\\
{\hat x}_{\mathrm{max}}  = f \cot  \langle \mathord{\buildrel{\lower3pt\hbox{$\scriptscriptstyle\rightharpoonup$}} 	\over {R}}, \mathord{\buildrel{\lower3pt\hbox{$\scriptscriptstyle\rightharpoonup$}} 	\over V} \rangle  = f\cot (\frac{{\pi  - \alpha }}{2}),
\end{array}  \right.
\end{equation}
where $ \alpha \in [0,\pi]$ is the given field-of-view angle of the horizontal-view camera as indicated in Fig.~\ref{fig:top_view}(a). From Eq.~(\ref{eq:xlr}), we have ${\hat x}_{\mathrm{max}} =-{\hat x}_{\mathrm{min}} >0$.

To enable the matching to the vector representation from the horizontal view, we further normalize the value range of ${\hat x}$ to $[-1, 1]$, i.e.,
\begin{equation} \label{eq:norm_top}
\left\{ \begin{array}{l}
x^{\mathrm{top}}= \dfrac{\hat x}{{f \cot (\frac{{\pi  - \alpha }}{2})}} \\
y^{\mathrm{top}}={\hat y}.
\end{array} \right.
\end{equation}
With this normalization, we actually do not need the actual value of $f$ in the proposed method. 

Let $O_k^{{{\rm top}} }$, ${k} \in {\mathcal{K}} \subset \{1,2,\cdots, N\}$ be the subset of detected subjects in the field of view in the top-view image. We can find the vector representation for all of them and sort them
in terms of $x^{\mathrm{top}}$ values in an ascending order. We then stack them together as 
\begin{equation} \label{eq:v_top}
{{\bf{V}}^{\rm top}}{\rm{ = (}}{{\bf{x}}^{{\mathop{\rm top}\nolimits} }} , {{\bf{y}}^{{\mathop{\rm top}\nolimits} }}) \in \mathbb{R} {^{{|{\mathcal{K}}|} \times 2}}
\end{equation}
where  ${|\mathcal{K}|}$ is the size of ${\mathcal{K}}$,  and ${{\bf{x}}^{{\mathop{\rm top}\nolimits} }}$ and  ${{\bf{y}}^{{\mathop{\rm top}\nolimits} }}$ are the column-wise vectors of all the $x^{\mathrm{top}}$ and $y^{\mathrm{top}}$ values of the subjects in the field of view, respectively. 

\subsubsection{Horizontal-View Vector Representation}
\label{sec:Hor_view_vec}

For each subject in the horizontal-view image, we also compute a vector representation to make it consistent to the top-view vector representation, i.e., $x$-value reflects the distance to the horizontal-view camera along the right direction and $y$-value reflects the depth to the horizontal-view camera. 
As shown in Fig.~\ref{fig:top_view}~(b), in the horizontal-view image, let $Q=(\tilde{x}, \tilde{y})$ and $h$ be the location and height of a detected subject, respectively. 
If we take the top-left corner of the image as the origin of the coordinate, $\tilde{x}-\frac{W}{2}$, with $W$ being the width of the horizontal-view image, is actually the subject's distance to the 
horizontal-view camera along the right direction.
To facilitate the matching to the top-view vectors, we normalize its value range to $[-1, 1]$ by
\begin{equation} \label{eq:norm_hor}
\left\{ \begin{array}{l}
x^{\mathrm{hor}}= \dfrac{{\tilde x} - \frac{W}{2}}{\frac{W}{2}} \\
y^{\mathrm{hor}}=\frac{1}{h},
\end{array} \right.
\end{equation}
where we simply take the inverse of the subject height as its depth to the horizontal-view camera. 

For all $M$ detected subjects in the horizontal-view image, we can find their vector representations and sort them
in terms of $x^{\mathrm{hor}}$ values in an ascending order. We then stack them together as 
 \begin{equation} \label{eq:v_hor}
 {{\bf{V}}^{\rm hor}}{\rm{ = (}}{{\bf{x}}^{{\mathop{\rm hor}\nolimits} }} , {{\bf{y}}^{{\mathop{\rm hor}\nolimits} }}) \in \mathbb{R} {^{{M} \times 2}}
 \end{equation}
 where  ${{\bf{x}}^{{\mathop{\rm hor}\nolimits} }}$ and  ${{\bf{y}}^{{\mathop{\rm hor}\nolimits} }}$ are the column-wise vectors of all the $x^{\mathrm{hor}}$ and $y^{\mathrm{hor}}$ values of the $M$ subjects detected in the  horizontal-view image, respectively. 

\subsection{Vector Matching}
\label{sec:vectormatch}

In this section we associate the subjects across two views by matching the vectors between the two vector sets ${{\bf{V}}^{{\mathop{\rm top}\nolimits} }}$ and ${{\bf{V}}^{{\mathop{\rm hor}\nolimits}}}$. Since the $x$ values of both vector sets have been normalized to the range of $[-1, 1]$, they can be directly compared. However, the $y$ values in these two vector sets are not comparable, although both of them reflect the depth to the horizontal-view camera: $y^{\mathrm{top}}$ values are in terms of number of pixels in the top-view image while $y^{\mathrm{hor}}$ values are in terms of the number of pixels in the horizontal-view image.
It is non-trivial to normalize them into a same scale given their errors in reflecting the true depth -- it is a very rough depth estimation by using $y^{\mathrm{hor}}$ since it is very sensitive to subject detection errors and height difference among subjects. 

We first find reliable subset matchings between ${{\bf{x}}^{{\mathop{\rm top}\nolimits} }}$ and ${{\bf{x}}^{{\mathop{\rm hor}\nolimits} }}$  and use them to estimate the scale difference between their corresponding $y$ values. More specifically, we find a scaling factor $\mu$ to scale $y^{\mathrm{top}}$ values to make them comparable to the $y^{\mathrm{hor}}$ values. For this purpose, we use a RANSAC-alike strategy~\cite{FISCHLER1981Random}: for each element $x^{\mathrm{top}}$ in ${{\bf{V}}^{{\mathop{\rm top}\nolimits} }}$, we find the nearest $x^{\mathrm{hor}}$ in ${{\bf{V}}^{{\mathop{\rm hor}\nolimits} }}$. If $|x^{\mathrm{top}}-x^{\mathrm{hor}}|$ is less than a very small threshold value,
we consider $x^{\mathrm{top}}$ and $x^{\mathrm{hor}}$ a matched pair and take the ratio of their corresponding $y$ values and the average of this ratio over all the matched pairs is taken as the scaling factor $\mu$. 

With the scaling factor $\mu$, we match ${{\bf{V}}^{{\mathop{\rm top}\nolimits} }}$ and ${{\bf{V}}^{{\mathop{\rm hor}\nolimits}}}$ 
using dynamic programming (DP)~\cite{Sniedovich2011Dynamic}. Specifically, we define a dissimilarity matrix ${{\bf D}}$ of dimension 
	$|{\mathcal{K}}|\times M$, where $D_{ij}$ is the dissimilarity between ${{\bf{V}}_i^{{\mathop{\rm top}\nolimits} }}$ and ${{\bf{V}}_j^{{\mathop{\rm hor}\nolimits}}}$ and it is defined by
	\begin{equation} \label{eq:dis}
D_{ij} 	= \lambda |x_i^{\mathrm{top}}-x_j^{\mathrm{hor}}|+|\mu y_i^{\mathrm{top}}-y_j^{\mathrm{hor}}|,
	\end{equation}
where $\lambda>0$ is a balance factor. Given that ${{\bf{x}}^{{\mathop{\rm top}\nolimits} }}$ and ${{\bf{x}}^{{\mathop{\rm hor}\nolimits} }}$ are both ascending sequences, we use dynamic programming algorithm to search a monotonic path in $\bf D$ from $D_{1,1}$ to $D_{|{\mathcal{K}}|,M}$ to build the matching between ${{\bf{V}}^{{\mathop{\rm top}\nolimits} }}$ and ${{\bf{V}}^{{\mathop{\rm hor}\nolimits}}}$
with minimum total dissimilarities. 
If a vector  ${{\bf{V}}^{{\mathop{\rm top}\nolimits} }}$ matches to multiple  vectors in  ${{\bf{V}}^{{\mathop{\rm hor}\nolimits}}}$, we only keep one with the smallest dissimilarity given in Eq.~(\ref{eq:dis}). After that, we check if a vector  ${{\bf{V}}^{{\mathop{\rm hor}\nolimits} }}$ matches to multiple  vectors in  ${{\bf{V}}^{{\mathop{\rm top}\nolimits}}}$ and we keep one with the smallest dissimilarity. These two-step operations will guarantee the resulting matching is one-on-one and we denote $\gamma$ to be the number of final matched pairs. Denote the resulting matched vector subsets to be ${{\bf{V}}_*^{{\mathop{\rm top}\nolimits} }}=({{\bf{x}}_*^{{\mathop{\rm top}\nolimits} }}, {{\bf{y}}_*^{{\mathop{\rm top}\nolimits} }})$ and ${{\bf{V}}_*^{{\mathop{\rm hor}\nolimits} }}=({{\bf{x}}_*^{{\mathop{\rm hor}\nolimits} }}, {{\bf{y}}_*^{{\mathop{\rm hor}\nolimits} }})$, both of dimension $\gamma\times 2$.  We define a matching cost between ${{\bf{V}}^{{\mathop{\rm top}\nolimits} }}$ and ${{\bf{V}}^{{\mathop{\rm hor}\nolimits}}}$ as
\begin{equation} \label{eq:cost}
\phi({{\bf{V}}^{{\mathop{\rm top}\nolimits} }}, {{\bf{V}}^{{\mathop{\rm hor}\nolimits} }}) 
=  \frac{1}{\gamma}\rho ^{\frac{L}{\gamma}}(\lambda \|{{\bf{x}}_*^{{\mathop{\rm top}\nolimits} }} -{{\bf{x}}_*^{{\mathop{\rm hor}\nolimits} }}\|_1 +\|{{\mu \bf{y}}_*^{{\mathop{\rm top}\nolimits} }} -{{\bf{y}}_*^{{\mathop{\rm hor}\nolimits} }}\|_1),
\end{equation}
where $\rho>1$ is a pre-specified factor and $L = \max(|{\mathcal{K}}|,M)$. In this matching cost, the term $\rho ^{\frac{L}{\gamma}}$ encourages the inclusion of more vector pairs into the final matching, which
is important when we use this matching cost to search for optimal camera location $O$ and view angle $\theta$ to be discussed next.

\subsection{Detecting Horizontal-View Camera and View Angle}
\label{sec:view_ang}

In calculating the matching cost of Eq.~(\ref{eq:cost}), we need to know the horizontal-view camera location $O$ and its view angle $\theta$ to compute the vector ${{\bf{V}}^{{\mathop{\rm top}\nolimits} }}$. In practice, we do not know  $O$ and $\theta$ priorly.  As mentioned earlier, we exhaustively try all possible values for $O$ and $\theta$ and then select the ones that lead to the minimum matching cost $\phi$. The matching with such minimum cost provides us the final cross-view subject association. For view angle $\theta$, we sample the its range $[0, 2\pi)$ uniformly with an interval of $\Delta \theta$ and in the experiments, we will report results by using different sample intervals. For the horizontal-view camera location $O$, we simply try every subject detected in the top-view image as the camera (wearer) location.  

\begin{figure}[htbp]
	\centering
	\includegraphics[width=\columnwidth]{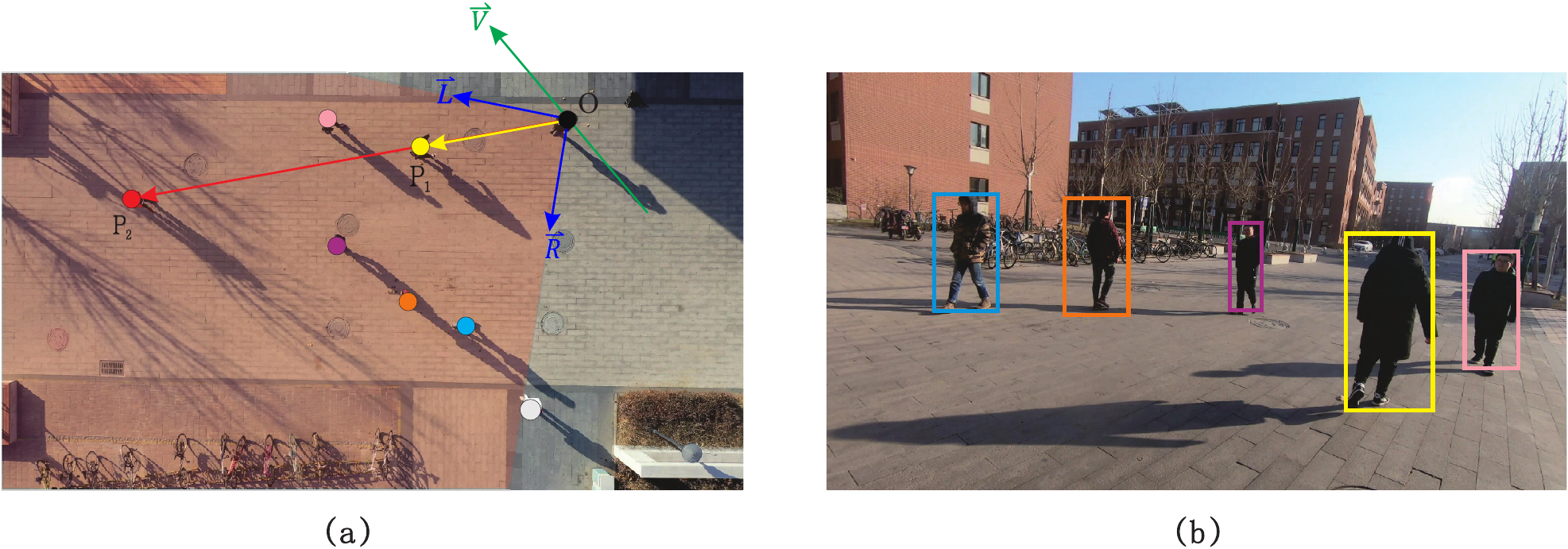}
	\caption{An illustration of mutual occlusion in the horizontal view. (a) Top-view image and (b) horizontal-view image.}
	\label{fig:occ_search}
\end{figure}

An occlusion in the horizontal-view image indicates that two subjects and the horizontal-view camera are collinear, as shown by $P_1$ and $P_2$ in Fig.~\ref{fig:occ_search}(a). 
In this case, the subject with larger depth, i.e., $P_2$,  is not visible in the horizontal view and we simply ignore this occluded subject in vector representation of ${{\bf{V}}^{{\mathop{\rm top}\nolimits} }}$.
In practice, we set a tolerance threshold $\beta=2^\circ$ and if $\langle \mathord{\buildrel{\lower3pt\hbox{$\scriptscriptstyle \rightharpoonup $}} \over {OP_1}}, \mathord{\buildrel{\lower3pt\hbox{$\scriptscriptstyle \rightharpoonup $}}  \over {OP_2}} \rangle<\beta$, we ignore the one with larger depth. The entire cross-view subject association algorithm is summarized in Algorithm~\ref{alg:practical}.
	
\begin{algorithm}[htbp]	
	\small
	\label{alg:practical}
	\caption{Cross-View Subject Association:}
	\KwIn{${\mathcal{T}}$, ${\mathcal{H}}$: Subjects detected in top view and horizontal view respectively; parameters ${\rho}$, ${\lambda}$.}
	\KwOut{Matched vector pair ${{\bf{V}}_*^{{\mathop{\rm top}\nolimits} }}$ and ${{\bf{V}}_*^{{\mathop{\rm hor}\nolimits} }}$; camera location $O^*$; camera-view angle $\theta^*$.}
	
	Compute the horizontal-view vector ${\bf{V}^{\rm hor}}$ by Eq.~$(\ref{eq:v_hor})$;
	
	\For{$O\in{\mathcal{T}}$} {
		\For{$\theta\in[0, 2\pi), \mathrm{with~step~length}~\Delta\theta$}{
			Compute the top-view vector ${\bf{V}^{\rm top}}$ by Eq.~$(\ref{eq:v_top})$;
			
			Estimate scaling $\mu$ as discussed in Section~\ref{sec:vectormatch}.
			
			Calculate $\bf{D}$ by Eq.~$(\ref{eq:dis})$ using $\mu$ and $\lambda$;
			
			Calculate ${{\bf{V}}^{{\mathop{\rm top}\nolimits} }}$, ${{\bf{V}}^{{\mathop{\rm hor}\nolimits} }}$ based on $\bf{D}$ by DP algorithm;
			
			Calculate $\phi$ by Eq.~$(\ref{eq:cost})$
		}
		Find $\theta$ with the minimum $\phi$;
	}
	\Return $O^*$,  $\theta^*$, ${{\bf{V}}_*^{{\mathop{\rm top}\nolimits} }}$ and ${{\bf{V}}_*^{{\mathop{\rm hor}\nolimits} }}$ with the minimum $\phi$.
\end{algorithm}

\section{Experiment}
\label{sec:experiment}

In this section, we first describe the dataset used for performance evaluation and then introduce our experimental results.

\subsection{ Test Dataset}

We do not find publicly available dataset with corresponding top-view and horizontal-view images/videos and ground-truth labeling of the cross-view subject association. 
Therefore, we collect a new dataset for performance evaluation. 
Specifically, we use a GoPro HERO7 camera (mounted over wearer's head) to take horizontal-view videos and a DJI ``yu" Mavic 2 drone to take top-view videos. 
Both cameras were set to have the same fps of 30. 
We manually synchronize these videos such that corresponding frames between them are taken at the same time.
We then temporally sample these two videos uniformly to construct frame (image) pairs for our dataset. 
Videos are taken at three different sites with different background and the sampling interval is set to 100 frames to ensure the variety of the collected images. 
Finally, we obtain 220 image pairs from top and horizontal views, and for both views, the image resolution is $2,688 \times 1,512$.
We label the same persons across two videos on all 220 image pairs.  
Note that, this manual labeling is quite labor intensive given the difficulty in identifying persons in the top-view images (see Fig.~\ref{fig:example} for an example). 

For evaluating the proposed method more comprehensively, we examine all 220 image pairs and consider the following five attributes: \textbf{Occ}: horizontal-view images containing partially or fully occluded subjects;
\textbf{Hor\_mov}: the horizontal-view images sampled from videos when the camera-wearer moves and rotates his head. 
\textbf{Hor\_rot}: the horizontal-view images sampled from videos when the camera-wearer rotates his head. 
\textbf{Hor\_sta}: the horizontal-view images sampled from videos when the camera-wearer stays static. 
\textbf{TV\_var}: the top-view images sampled from videos when the drone moves up, down and/or change camera-view direction. 
Table~\ref{tab:data} shows the number of image pairs with these five attributes, respectively.  Note that some image pairs show multiple attributes listed above.

\begin{table}[htbp] 
	\caption{Number of image pairs with the considered attributes.}
	\label{tab:data}
	\footnotesize
	\centering
		\vspace{5 pt}
	\begin{spacing}{1.25}
		\begin{tabular}{lccccc}	
			\hline
			Attribute  & \textbf{Occ} & \textbf{Hor\_mov} & \textbf{Hor\_rot} & \textbf{Hor\_sta} & \textbf{TV\_var} \\
			\# image pairs   & 96  & 62       & 124       & 96       & 30      \\ \hline		
		\end{tabular}
	\end{spacing}
\end{table}

\begin{figure}
	\centering
	\includegraphics[width=0.5\textwidth]{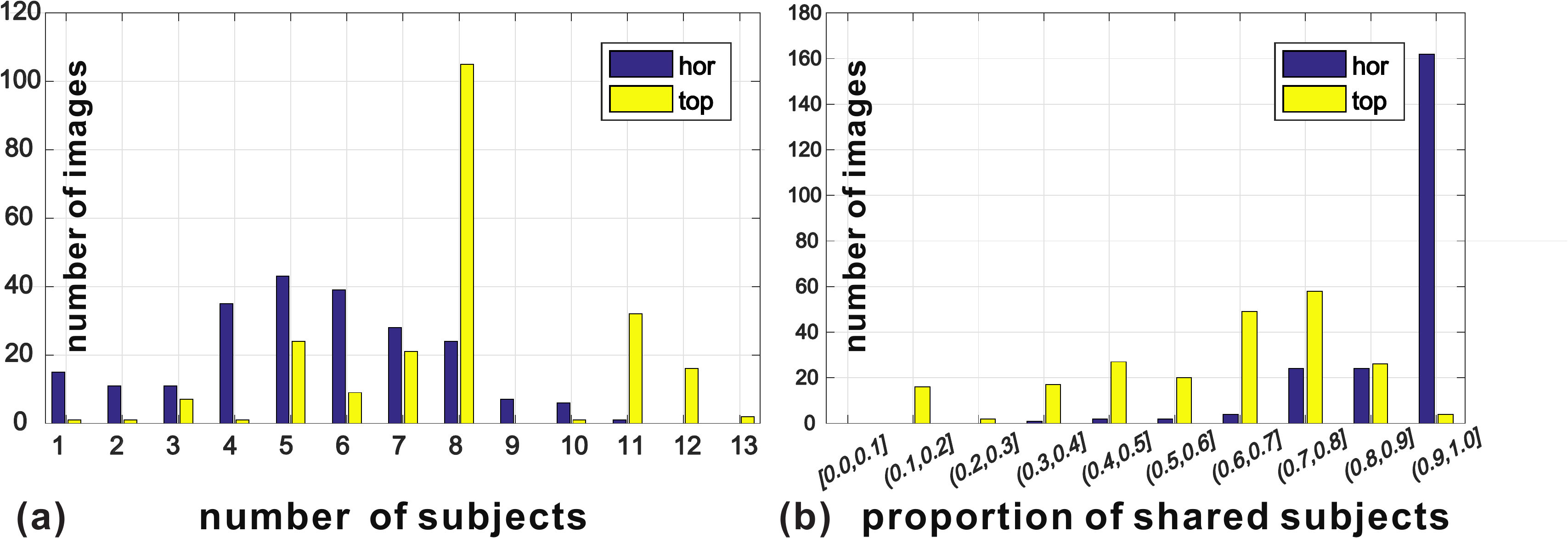}
	\caption{(a) Histogram of the crowdedness in top- and horizontal-view images, respectively. (b) Histogram of the proportion of the shared subjects in top- and horizontal-view images, respectively.}
	\label{fig:data_set}
\end{figure}

For each pair of images, we analyze two more properties. One is the number of subjects in an image, which reflects the level of crowdedness. The other is the proportion between the number of shared subjects in two views and the total number of subjects in an image. Both of them can be computed against either the top-view image or the horizontal-view image and their histograms on all 220 image pairs 
are shown in Fig.~\ref{fig:data_set}.

In this paper, we use two metrics for performance evaluation. 1) The accuracy in identifying the horizontal-view camera wearer in the top-view image,  and 2) 
the precision and recall of cross-view subject association. 
We do not include the camera-view angle $\theta$ for evaluation because it is difficult to annotate its ground truth. 


\subsection{Experiment Setup}

We implement the proposed method in Matlab and run on a desktop computer with an Intel Core i7 3.4GHz CPU. 
We use the general YOLO~\cite{Redmon2016You} detector to detect subjects in the form of bounding boxes in both top-view and horizontal-view images~\footnote{We use the YOLOv3 version detector. For top-view subject detection, we fine-tune the network using 600 top-view human images that have no overlap with our test images.}. 
The pre-specified parameters $\rho$ and $\lambda$ are set to 25 and 0.015 respectively. 
We will further discuss the influence of these parameters in Section~\ref{sec:Analysis}.

We did not find available methods with code that can directly handle our top- and horizontal-view subject association.  One related work is~\cite{ardeshir2018integrating} for cross-view matching. However, we could not include it directly into comparison because 1) its code is not available to public, and 2) it computes optical flow for $\theta$ and therefore cannot handle a pair of static images in our dataset. Actually, the method in~\cite{ardeshir2018integrating} assumes a certain slope view angle of the top-view camera and use appearance matching for cross association. This is similar to the appearance-matching-based person re-id methods. 

In this paper, we chose a recent person re-id method~\cite{sun2018beyond} for comparison. We take each subject detected in the horizontal-view image as query and search it in the set of subjects detected in the top-view image. 
We tried two versions of this re-id method: one is retrained from scratch using 1,000 sample subjects collected by ourselves (no overlap with the subjects in our test dataset) and the other is to fine-tune from the version provided in~\cite{sun2018beyond} these 1,000 sample subjects.

\subsection{Results} 

\begin{figure}[htbp]
	\centering
	\includegraphics[width=0.45\textwidth]{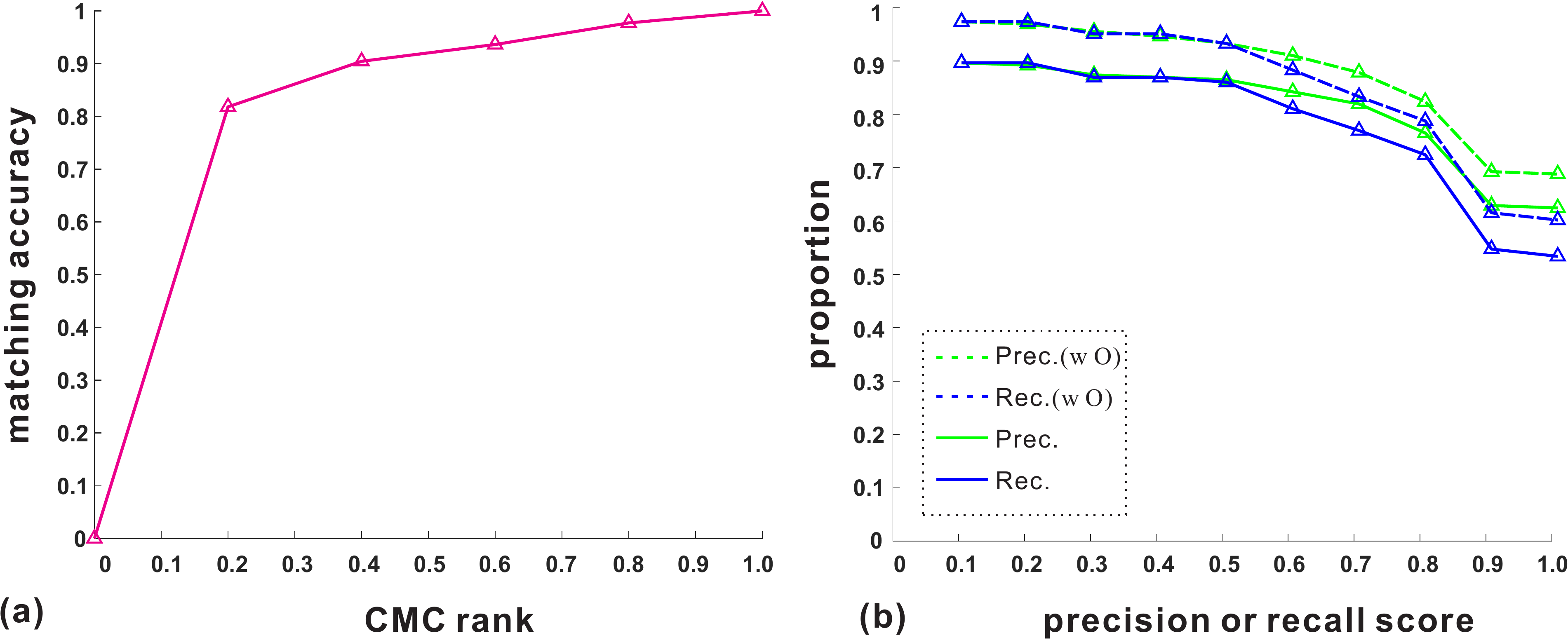}
	\caption{(a) The CMC curve for horizontal-view camera detection. (b) Precision and recall scores in association, 
		where the horizontal axis denotes a precision or recall score $x$, and the vertical coordinate denotes the proportion of image pairs with corresponding precision or recall score that is greater than $x$.}
	\label{fig:rank_curve}
\end{figure}

We apply the proposed method to all 220 pairs of images in our dataset. We detect the horizontal-view camera wearer on the top-view image as described in \ref{sec:view_ang} and the detection accuracy is 
84.1\%. We also use the Cumulative Matching Characteristic (CMC) curve to evaluate the matching accuracy, as shown in Fig.~\ref{fig:rank_curve}(a), where the
horizontal and vertical axes are the CMC rank and the matching accuracy respectively.

For a pair of images, we use the precision and recall scores to evaluate the cross-view subject association. 
As shown in Table \ref{tab:res}, the average precision and recall scores of our method are 79.6\% and 77.0\% respectively. 
In this table, `Ours w $O$' indicates the use of our method by giving the ground-truth camera location $O$.
We can find in this table that the re-id method, either retrained or fune-tuned, produces very poor result, 
which confirms the difficulty in using appearance features for the proposed cross-view subject association. 

\begin{table}[htbp]
	\caption{Comparative results of different methods. Prec.Avg and Reca.Avg are the average precision and recall scores over all image pairs. Prec.@1 and Reca.@1 are the proportion of image pairs with precision and recall scores of 1, respectively.}
	\vspace{-0.2cm}
	\label{tab:res}
	\centering
	\footnotesize
	\begin{spacing}{1.25}
		\begin{center}
			\begin{tabular*}{0.475\textwidth}[c]{lccccc}
				\Xhline{1pt}
				Method &Prec.Avg	&Reca.Avg 	&Prec.@1	 &Reca.@1 	 \\  	\hline						
				Re-id~(fine-tune)  &14.0 &16.0 &- &-\\
				Re-id~(retrain) &22.0 &24.0 &- &- \\
				Ours (w $O$)  &86.6 &84.2 &66.4 &57.7\\
				Ours &79.6 &77.0 &60.0 &50.9 \\
				\hline			
			\end{tabular*}
		\end{center}
	\end{spacing}
\end{table}

We also calculate the proportion of all the image pairs with precision or recall score of 1 (Prec.@1 and Reca.@1). They reach 60.0\% and 50.9\% respectively.
The distributions of these two scores on all 220 image pairs are shown in Fig.~\ref{fig:rank_curve}(b).
In Table~\ref{tab:att_per}, we report the evaluation results on different subsets with respective attributes. 
We can see that the proposed method is not sensitive to the motion of both top-view and horizontal-view cameras, which is highly desirable for motion-camera applications. 

\begin{table}[htbp] 
	\caption{Comparative results on the subsets with different attributes.}
	\label{tab:att_per}
	\centering
	\footnotesize
	\begin{spacing}{1.25}
		\begin{tabular*}{0.45\textwidth}{lccccc}			
			\Xhline{1pt}
			Attr  & \textbf{Occ} & \textbf{Hor\_mov} & \textbf{Hor\_rot} & \textbf{Hor\_sta} & \textbf{TV\_var} \\ \hline
			Prec.Avg   &76.6   &78.3        &80.5        &78.4       &53.3      \\ 
			Rec.Avg   &74.5   &74.9       &77.9      &75.7       &53.3     \\ \hline
		\end{tabular*}
	\end{spacing}
\end{table}

\subsection{Ablation Studies}

\label{sec:Analysis}

\textbf{Step Length for $\theta$.}
We study the influence of the value $\Delta \theta$, the step length for searching optimal camera view angle $\theta$ in the range $[0, 2\pi)$. We set the value of $\Delta\theta$ to 1$^\circ$, 5$^\circ$ and 10$^\circ$, respectively and the association results are shown in Table~\ref{tab:para-2}. 
As expected, $\Delta\theta=1^\circ$ leads to the highest performance, although a larger step length, such as $\Delta\theta=5^\circ$ also produces acceptable results.

\begin{table}[htbp]
	\caption{Results by using different values for $\Delta \theta$.}
	\vspace{-0.cm}
	\label{tab:para-2}
	\centering
	\footnotesize
	\begin{spacing}{1.25}
		\begin{center}
			\begin{tabular*}{0.45\textwidth}[c]{lccccc}
				\Xhline{1pt}
				Step length &Prec.Avg	 &Reca.Avg 	&Prec.@1	 &Reca.@1 	 \\  	\hline						
				$\Delta \theta$= 1\degree &\textbf{79.6} &\textbf{77.0} & \textbf{60.0} &\textbf{50.9} \\
				$\Delta \theta$= 5\degree &78.8 &76.9 &59.6 &51.8 \\
				$\Delta \theta$ = 10\degree &72.5 &71.1 &53.2 &46.8 \\
				
				\hline			
			\end{tabular*}
		\end{center}
	\end{spacing}
\end{table}

\textbf{Vector representation.}
Next we compare the association results using different vector representation methods as shown in Table~\ref{tab:vec_rep}. The first row denotes that we represent the subjects in two views by one-dimensional vectors $\bf{x}^\mathrm{top}$ and $\bf{x}^\mathrm{hor}$ respectively. 
The second row denotes that we represent the subjects in two views by one-dimensional vectors $\bf{y}^\mathrm{top}$ and $\bf{y}^\mathrm{hor}$, respectively, which are simply normalized to the range $[0,1]$ to make them comparable. 
The third row denotes that we combine the one-dimensional vectors for the first and second rows to represent each view, which differs from our proposed method (the fourth row of Table~\ref{tab:vec_rep}) only on the normalization of $\bf{y}^\mathrm{top}$ and $\bf{y}^\mathrm{hor}$ -- our proposed method uses a RANSAC strategy.
By comparing the results in the third and fourth rows, we can see that the use of RANSAC strategy for estimating the scaling factor $\mu$ does improve the final association performance. 
The results in the first and second rows show that using only one dimension of the proposed vector representation cannot achieve performance as good as the proposed method that combines both dimensions. 
We can also see that $\bf{x}^\mathrm{top}$ and $\bf{x}^\mathrm{hor}$  provides more accurate information than  $\bf{y}^\mathrm{top}$ and $\bf{y}^\mathrm{hor}$ when used for cross-view subject association. 

\begin{table}[htbp]
	\caption{Comparative study of using different vector representations. $\bf{\bar y}^\mathrm{top}$ and $\bf{\bar y}^\mathrm{hor}$ are normalized results of $\bf{y}^\mathrm{top}$ and $\bf{y}^\mathrm{hor}$, respectively, by simply scaled to $[0,1]$.}
	\vspace{-0.2cm}
	\label{tab:vec_rep}
	\centering
	\footnotesize
	\begin{spacing}{1.25}
		\begin{center}
			\begin{tabular*}{0.475\textwidth}[c]{lccccc}
				\Xhline{1pt}
				Vector	&Prec.Avg	 &Reca.Avg 	&Prec.@1	 &Reca.@1 	 \\  	\hline						
				$\bf{x}^\mathrm{top,hor}$  &63.2 &61.6 &41.8 &35.9\\
				$\bf{\bar y}^\mathrm{top,hor}$ &23.4 &13.4 &6.8 &0.9 \\
				$\bf{x}^\mathrm{top,hor}$,  $\bf{\bar y}^\mathrm{top,hor}$   &67.7 &66.6 &46.4 &42.7 \\
				Ours &\textbf{79.6} &\textbf{77.0} & \textbf{60.0} &\textbf{50.9} 	\\  
				\hline			
			\end{tabular*}
		\end{center}
	\end{spacing}
\end{table}

\textbf{Parameters selection.}
There are two free parameters $\rho$ and $\lambda$ in Eq.~(\ref{eq:cost}). We select different values for them and see their influence to the final association performance. 
Table~\ref{tab:para} reports the results by varying one of these two parameters while fixing the other one.
We can see that the final association precision and recall scores are not very sensitive to the selected values of these two parameters. 

\begin{table}[htbp]
	\caption{Results by varying values of $\rho$ and $\lambda$.}
	\vspace{-0.2cm}
	\label{tab:para}
	\centering
	\footnotesize
	\begin{spacing}{1.25}
		\begin{center}
			\begin{tabular*}{0.5\textwidth}[c]{lcc|lcc}
				\Xhline{1pt}
				$\rho$	&Prec.Avg	 &Reca.Avg &$\lambda$	&Prec.Avg	  &Reca.Avg	 \\  	\hline
				$\rho=10$  &76.4 &73.7	
				&$\lambda = 0.005$ &79.2 &76.6\\						
				$\rho=25$ &\textbf{79.6} &\textbf{77.0}  & $\lambda = 0.015$ &\textbf{79.6} &\textbf{77.0} \\
				$\rho=40$  &79.1 &76.7 & $\lambda = 0.025$ &78.4 &75.9\\
				
				\Xhline{1pt}							
			\end{tabular*}
		\end{center}
	\end{spacing}
\end{table}

{\bf{Detection method.}}
In order to analyze the influence of subjects detection's accuracy to the proposed cross-view association, we tried the use of different subject detections. As shown in Table~\ref{tab:det},
in the first row, we use manually annotated bounding boxes of each subject on both views for the proposed association. 
In the second and third rows, we use manually annotated subjects on top-view images and horizontal-view images, respectively,  while using automatically detected subjects~\cite{Redmon2016You} 
on the other-view images. 
In the fourth row, we automatically detect subjects in both views first, and then only keep those that show an IoU$>0.5$ (Intersection over Union) against a manually annotated subject, in terms of their bounding boxes. 
We can see that the use of manually annotated subjects produces much better cross-view subject association. This indicates that further efforts on improving subject detection will benefit the association. 

\begin{table}[htbp]
	\caption{Results by using different methods for subject detection.}
	\vspace{-0.5cm}
	\label{tab:det}
	\centering
	\footnotesize
	\begin{spacing}{1.25}
		\begin{center}
			\begin{tabular*}{0.5\textwidth}[c]{lccccc}
				\Xhline{1pt}
				Subjects detection &Prec.Avg	 &Reca.Avg 	&Prec.@1	 &Reca.@1 	 \\  	\hline						
				Manual   &{83.5} &{80.5} &\textbf{76.8} &\textbf{61.4}\\
				Manual-Top   &\textbf{84.8} &\textbf{82.0} &70.5 &59.1 \\
				Manual-Hor   &80.6 &77.4 &69.1 &55.5 \\				
				Automatic w selection &80.7 &76.1 &69.6 &52.7 \\
				Automatic (Ours)  &79.6 &77.0 &60.0 &50.9\\
				\hline			
			\end{tabular*}
		\end{center}
	\end{spacing}
\end{table}


\begin{figure}[htbp]
	\centering
	\includegraphics[width=0.45\textwidth]{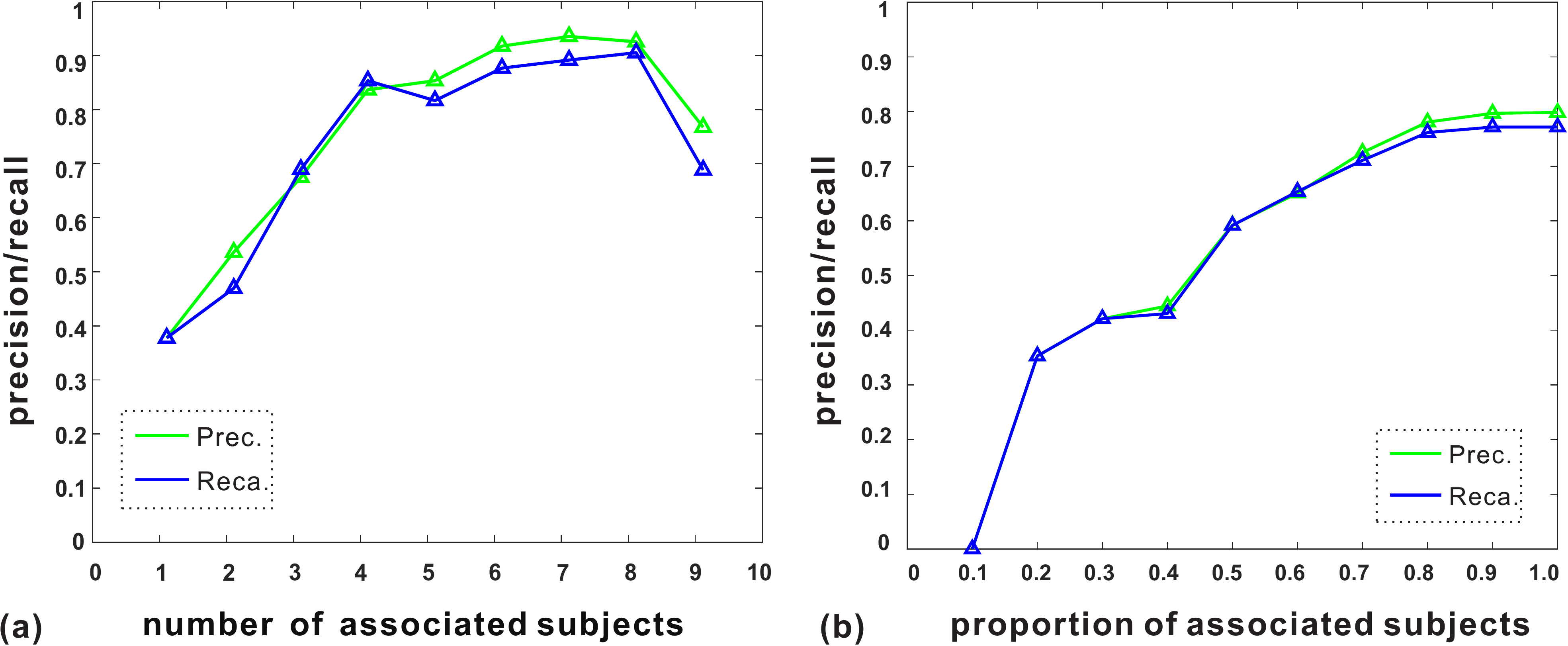}
	\caption{(a) Association performance for image pairs with different number of associated subjects. (b) Association performance for image pairs with different proportion of associated subjects.}
	\label{fig:pre_rec_num_scale}
\end{figure}

\begin{table}[htbp]
	\caption{Association results using the proposed method with and without handling occlusions.}
	\label{tab:occ}
	\centering
	\footnotesize
	\begin{spacing}{1.25}
		\begin{center}
			\begin{tabular*}{0.45\textwidth}[c]{ l  c c c c}
				\Xhline{1pt}
				\multirow{2}{*}{{Method}}
				&\multicolumn{2}{c}{Whole dataset} 	&\multicolumn{2}{c}{Occ subset}	  \\ \cline{2-5}
				&{Prec.Avg}	 &Reca.Avg		&Prec.Avg	  &Reca.Avg		\\  	\hline				
				Ours 	&79.6	 &77.0 		&76.6	 &74.5	  	\\
				Ours(w/o occ)    &65.2	 &65.1		&46.1	 &46.8	      \\ 	
				\Xhline{1pt}							
			\end{tabular*}
		\end{center}
	\end{spacing}\vspace{-0.5cm}
\end{table}

\begin{figure*}[htbp]
	\centering
	\includegraphics[width=0.95\textwidth]{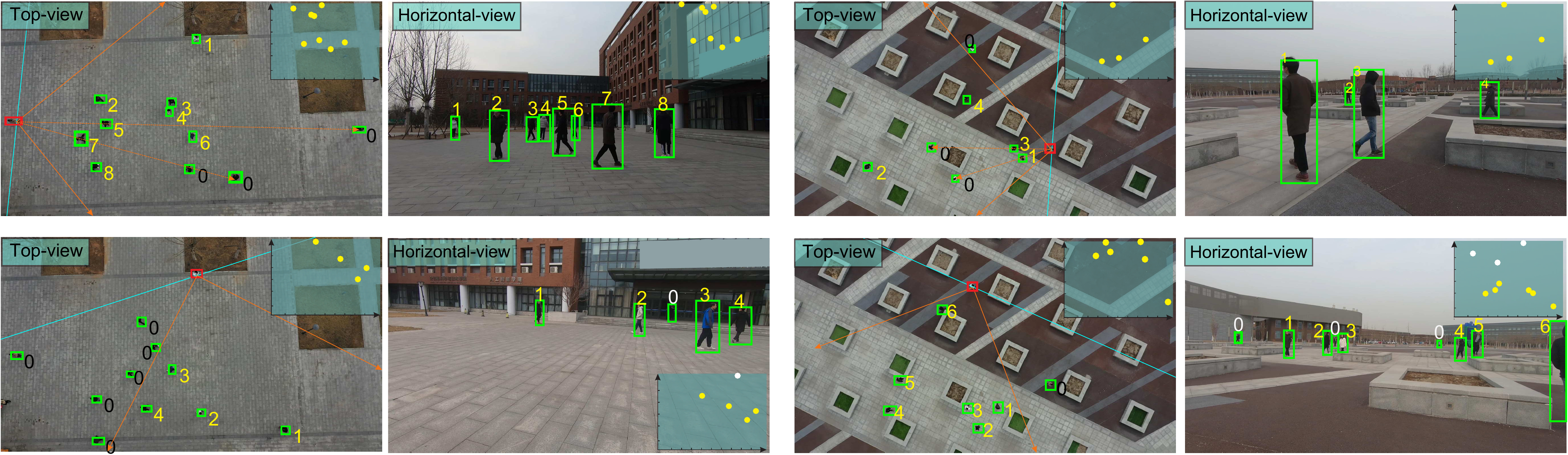}
	\caption{Row 1: Two sample results on image pairs with occlusions. Row 2: Two sample results with large number of unshared subjects between two views.
		Vector sets ${{\bf{V}}^{{\mathop{\rm top}\nolimits} }}$ and ${{\bf{V}}^{{\mathop{\rm hor}\nolimits}}}$ are shown in the top-right corner of every image.}
	\label{fig:dis_occ_det}
\end{figure*}

\begin{figure*}[htbp]
	\centering
	\includegraphics[width=0.95\textwidth]{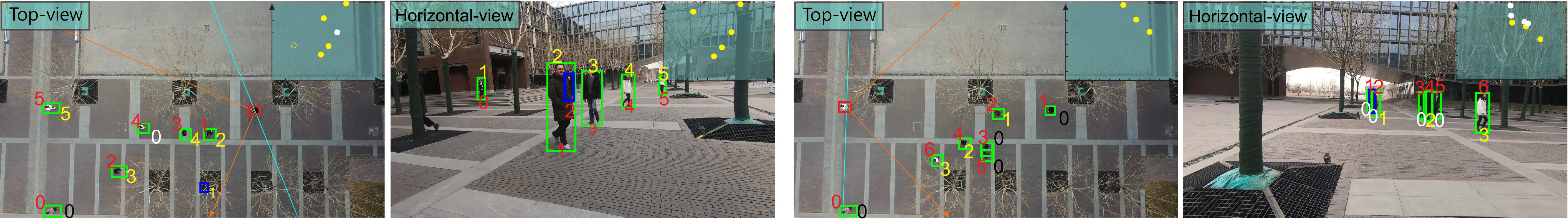}
	\caption{Two failure cases.}
	\label{fig:dis_fail}
\end{figure*}

\subsection{Discussion}

\textbf{Number of associated subjects.}
We investigate the correlation between the association performance and the number of associated subjects.
Figure~\ref{fig:pre_rec_num_scale}(a) shows the average association performance on the image pairs with different number of associated subjects. 
We can see that the association results get worse when the number of associated subjects is too high or too low. 
When there are too many associated subjects, the crowded subjects in the horizontal view may prevent the accurate detection of subjects.
When there are two few subjects, the constructed vector representation is not sufficiently discriminative to locate the camera location $O$ and camera-view angle $\theta$.
Figure~\ref{fig:pre_rec_num_scale}(b) shows the average association performance on the image pairs with different proportions of associated subjects.
More specifically, the performance at $x$ along the horizontal axis is the average precision/recall score on all the image pairs with the proportion of associated subjects (to
the total number of subjects in the top-view image) less than $x$.
This confirms that on the images with higher such proportion, the association can be more reliable.

\textbf{Occlusion.}
Occlusions are very common, as shown in  Table~\ref{tab:data}. Table~\ref{tab:occ} shows the association results on the entire dataset and the subset of data with occlusions, by using the proposed method with and without the step of identifying and ignoring occluded subjects. We can see that our simple strategy for handling occlusion can significantly improve the association performance on the image pairs with occlusions. 
Sample results on image pairs with occlusions are shown in the top row of Fig.~\ref{fig:dis_occ_det}, where associated subjects bear same number labels. We can see that occlusions occur more often when 1) the subjects are crowded, and 2) one subject is very close to the horizontal-view camera.

\textbf{Proportion of shared subjects.}
It is a common situation that many subjects in two views are not the same persons. In this case, the shared subjects may only count for a small proportion in both top- and horizontal-views.
Two examples are shown in the second row of Fig.~\ref{fig:dis_occ_det}. In the left, we show a case where many subjects in the top view are not in the field of view of the horizontal-view camera. 
In the right, we show a case where many subjects in the horizontal view are too far from the horizontal-view camera and not covered by the top-view camera.
We can see that the proposed method can handle these two cases very well, by exploring the spatial distribution of the shared subjects.

\textbf{Failure case.}
At last, we give two failure cases as shown in Fig.~\ref{fig:dis_fail} -- one caused by the error in subject detection (blue boxes) and the other is caused by the close distance of multiple subjects, e.g, subjects 3,4 and 5,
in either top or horizontal view, which lead to error detection of occlusions and incorrect vector representations.

\section{Conclusion}
\label{sec:conclusion}
In this paper, we developed a new method to associate multiple subjects across top-view and horizontal-view images by modeling and matching the subjects' spatial distributions. We constructed a vector representation for all the detected subjects in the horizontal-view image and another vector representation for all the detected subjects in the top-view image that are located in the field of view of the horizontal-view camera.  These two vector representations are then matched for cross-view subject association. We proposed a new matching cost function with which we can further optimize for the location and view angle of the horizontal-view camera in the top-view image. We collected a new dataset, as well as manually labeled ground-truth cross-view subject association, and experimental results on this dataset are very promising.

{\small
\bibliographystyle{ieee}
\bibliography{MOA}
}

\end{document}